\documentclass[sigconf]{acmart}

\AtBeginDocument{%
  \providecommand\BibTeX{{%
   \normalfont B\kern-0.5em{\scshape i\kern-0.25em b}\kern-0.8em\TeX}}}

\begin{document}

\title{Federated Learning: Opportunities and Challenges}

\author{Priyanka Mary Mammen}
\affiliation{%
  \institution{University of Massachusetts, Amherst}
  \streetaddress{1 Th{\o}rv{\"a}ld Circle}
 }
\email{pmammen@cs.umass.edu}


\begin{abstract}
 Federated Learning (FL) is a concept first introduced by Google in 2016, in which multiple devices collaboratively learn a machine learning model without sharing their private data under the supervision of a central server. This offers ample opportunities in critical domains such as healthcare, finance etc, where it is risky to share private user information to other organisations or devices. While FL appears to be a promising Machine Learning (ML) technique to keep the local data private, it is also vulnerable to attacks like other ML models. Given the growing interest in the FL domain, this report discusses the opportunities and challenges in federated learning.
\end{abstract}





\maketitle
\pagestyle{plain}
\section{Introduction}
Artificial Intelligence (AI)/Machine Learning (ML) started getting popular in the last few 4-5 years when AI beat humans in a board game named Alpha-Go\cite{silver2017mastering}. Availability of Big-data and powerful computing units further accelerated the adoption of Machine Learning technologies in domains such as finance, healthcare, transportation, customer services, e-commerce, smart home applications etc. With this widespread adoption of ML techniques, it is therefore important to ensure the security and privacy of the techniques. In most of the machine learning applications, data from various organizations or devices are aggregated in a central server or a cloud platform for training the model. This is a key limitation especially when the training data set contains sensitive information and therefore, poses security threats. For example, to develop a breast cancer detection model from MRI scans, different hospitals can share their data to develop a collaborated ML model. Whereas, sharing private patient information to a central server can reveal sensitive information to the public with several repercussions. In such scenarios, Federated Learning can be better option.Federated Learning is a collaborative learning technique among devices/organizations, where the model parameters from local models are shared and aggregated instead of sharing their local data. 
\par The notion of Federated Learning is introduced by Google in 2016, where they first applied  in google keyboard to collaboratively learn from several android phones \cite{mcmahan2017federated}. Given that FL can be applied to any edge device, it has the potential to revolutionize critical domains such as healthcare, transportation, finance, smart home etc. The most prominent example is when the researchers and medical practitioners from different parts of the world collaboratively developed an AI pandemic engine for COVID-19 diagnosis from chest scans \cite{weblink}. Another interesting application would be in transportation networks, for training the vehicles for autonomous driving and city route planning. Similarly for smart-home applications, edge devices in different homes can collaboratively learn on context aware policies using a federated learning framework \cite{yu2020learning}.
\par While the applications are many, there are several challenges associated with federated learning. The challenges can be broadly classified into two: training-related challenges and security challenges. Training related challenges encompass the communication overhead during multiple training iterations, heterogeneity of the devices participating in the learning and heterogeneity of data used for training .Whereas security challenges include the privacy and security threats due to the presence of adversaries ranging from malicious clients in the local device to a malicious user who has only a black-box access to the model.  In FL, although the private data does not leave the device, it might be still possible for an adversary or a curious observer to learn the presence of a data point used for training in the local models. In order to overcome this attack, some kind of cryptographic technique is required to keep the information differentially private \cite{geyer2017differentially}. Whereas security attacks can be mostly induced by  the presence of malicious clients in the learning, and they can be either targeted or non-targeted. In targeted attacks, the adversary wants to manipulate the labels on specific tasks. Whereas in non-targeted attacks, the motivation of the adversary is just to compromise the accuracy of the global model. The defense mechanisms require to detect malicious devices and remove them from further learning or nullify the effect on the global model induced by the malicious devices \cite{fang2020local}.

\par Numerous research efforts have been undertaken in the last few years to fortify the Federated Learning domain and it has some effect on the performance parameters like accuracy, computational costs etc. Being a distributed system,its difficult to identify malicious participants in FL. FL domain has grown so far from being a small application introduced by Google. Researchers have introduced various FL architectures, incentive mechanisms to foster user participation, cloud services for FL etc. Motivated by this growing interest in the Federated Learning domain, we present this survey paper. The recent works \cite{kairouz2019advances,yang2019federated,mothukuri2020survey,aledhari2020federated} are focused either on different federated learning architecture or on different challenges in FL domain. Whereas so many interesting advancements are taking place beyond applying it to new application domains and fostering the security aspects Therefore, in this survey we also try to cover the recent developments along with providing  give a general overview on Federated Learning applications and security concerns in the domain.

 \begin{figure}
    \centering
    \includegraphics[width=0.42\textwidth]{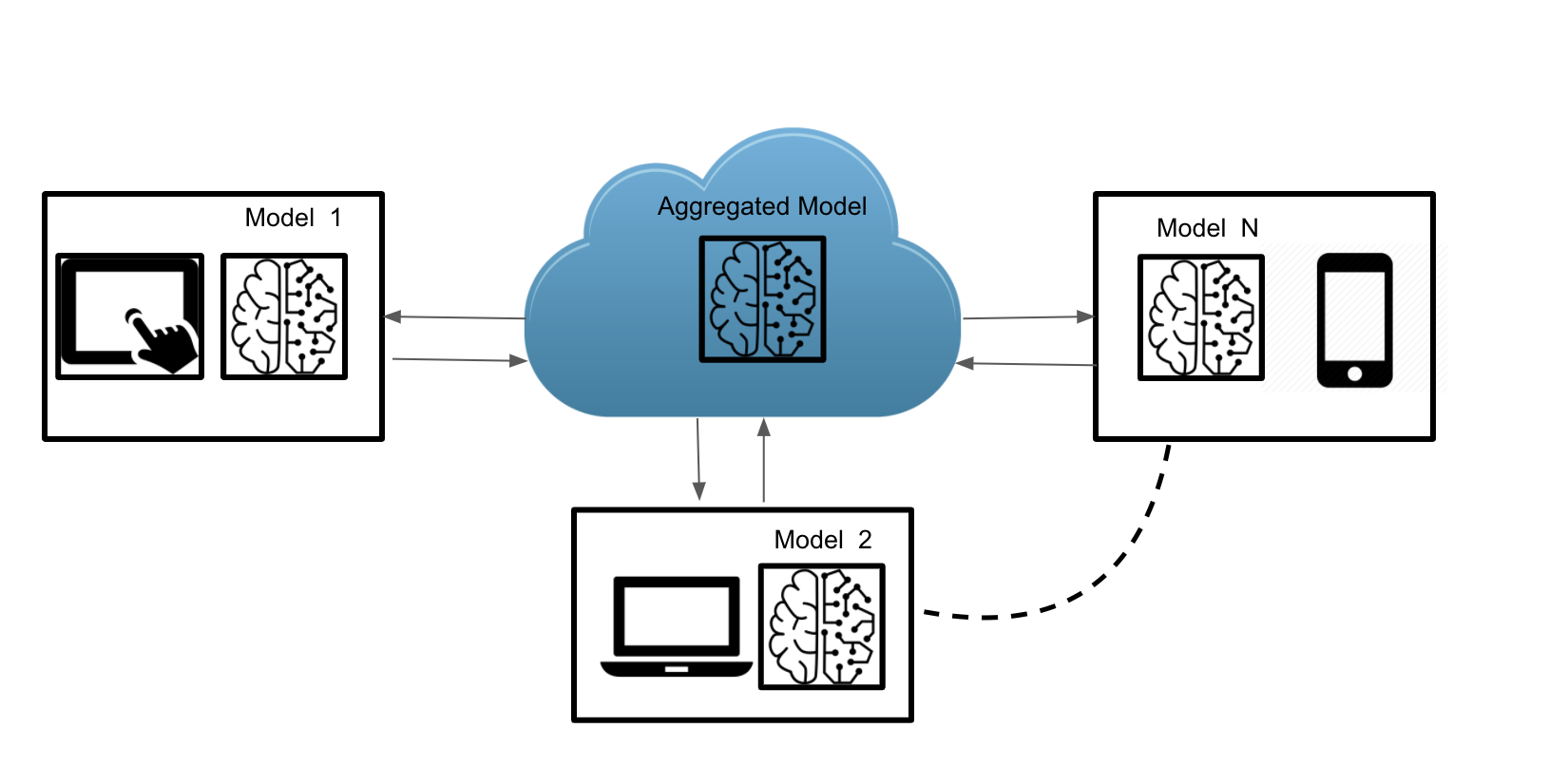}

  \caption{Overview of Federated Learning across devices.}
  \label{fig:perf1}

    \includegraphics[width=0.4\textwidth]{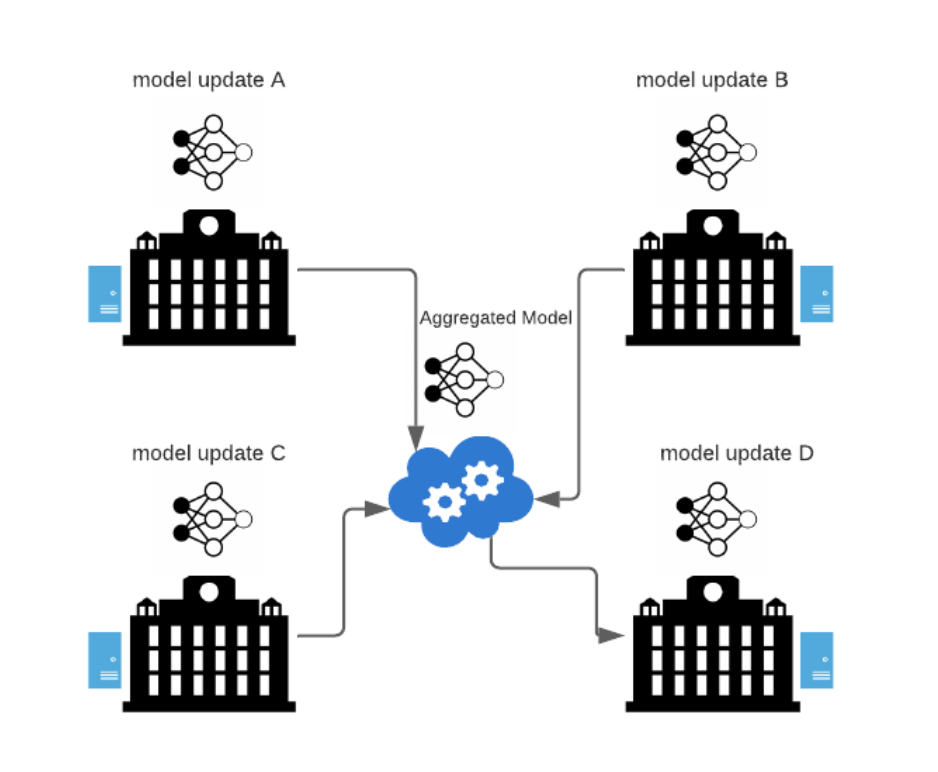}

  \caption{Overview of Federated Learning across organisations}
  \label{fig:perf2}
\end{figure}
\section{Overview of Federated Learning}

 A simple representation of Federated Learning is shown in fig. \ref{fig:perf1} and fig. \ref{fig:perf2}. Federated Learning has primarily four main steps:
 \begin{itemize}
     \item Client Selection/Sampling : Server either randomly picks desired participants from a pool of devices or use some algorithm for client selection. \cite{zhan2020learning,khan2020federated,yu2020sustainable} talk about some client selection techniques for FL.
     \item Parameter Broadcasting: Server broadcasts the global model parameters to the selected clients
     \item Local Model Training: The clients will parallelly retrain the models using their local data.
     \item Model Aggregation: Clients will send back their local model parameters to the server and model parameters will be aggregated towards the global model.
 \end{itemize}
 The above steps will be repeated in an iterative manner for n times as desired.
\section{Types of Federated Learning}
In this section, we introduce different types of Federated Learning frameworks.
\begin{itemize}
    \item Vertical Federated Learning -Vertical Federated Learning is used for cases in which each device contains dataset with different features but from sample instances. For instance, two organisations have data about the same group of people with different feature set can use Vertical FL to build a shared ML model. 
    \item Horizontal Federated Learning - Horizontal Federated Learning is used for cases in which each device contains dataset with the same feature space but with different sample instances. The first use case of FL- Google keyboard uses this type of learning in which the participating mobile phones have different training data with same features.
    \item Federated Transfer Learning - Federated Transfer learning is similar to the traditional Machine Learning, where we want to add a new feature on a pre-trained model. The best example would be for giving an extension to the vertical federated learning - If we want to extend the ML to more number of sample instances which are not present in all of the collaborating organisations.
    \item Cross-Silo Federated Learning - Cross- Silo Federated Learning is used when the participating devices are less in number and available for all rounds. The training data can be in horizontal or vertical FL format. Mostly cross-silo is used for cases with organisations. Works such as \cite{zhang2020batchcrypt} use cross-silo FL to develop their model.
    \item Cross-Device Federated Learning - Scenarios with a large number of participating devices use Cross-device Federated Learning. Client-selection and incentive designs\cite{yu2020sustainable} are some notable techniques needed to facilitate this type of FL. 
    
\end{itemize}
\section{Applications}
\subsection{Healthcare}

Electronic Health Records (EHR) is considered as the main source of healthcare data for machine learning applications \cite{ghassemi2020review}. If ML models are trained only using the limited data available in a single hospital, it might introduce some amount of bias in the predictions. Thus, to make the models more generalizable, it requires training with more data, which can be realized by sharing data among organizations. Given the sensitive nature of the healthcare data, it might not be feasible to share the electronic health records of patients among hospitals. In such situations, federated learning can serve as an option for building a collaborative learning model for healthcare data.
\subsection{Transportation}
With the increase in the ubiquity of sensors in vehicular networks, it is feasible to capture more data and train ML models. Machine Learning based models are generally applied to both vehicle management and traffic management \cite{tan2020federated}. The current autonomous driving decisions are limited by the dynamic nature of the surroundings as the training is carried out offline. FL can rescue such situations by online training vehicles from different geographical locations which can facilitate accurate labelling of the features. Similarly for traffic flow prediction techniques, a large amount of data is required, but most of the data is divided among various organizations and cannot be exchanged to protect the privacy \cite{liu2020privacy}. To address such situations also, we can deploy FL methods.
\subsection{Finance}
One best use of federated learning in finance is in the banking sector, for loan risk assessment \cite{cheng2020federated}. Normally banks use white-listing techniques to rule out the customers using their credit card reports from the central banks. Factors such as taxation, reputation etc can also be utilized for risk management by collaborating with other finance institutions and e-commerce companies. As it is risky to share private information of customers among organizations, they can make use of FL to build a risk assessment ML model. 

\subsection{Natural Language Processing}
Natural Language Processing (NLP) is one of the most common applications which is built on machine learning models. It helps us understand human language semantics in a better way. However, it requires huge amount of data to train highly accurate language models. This data can be easily gathered from mobile phones, tablets etc. Again, privacy comes as a bottleneck here for centralized language learning models, as the textual information from each edge device contains user information. In \cite{garcia2020decentralizing}, the authors have shown that it is feasible to build NLP models using a FL framework.

\section{Training Bottlenecks}
Being a distributed system, Federated Learning faces several challenges during the training time.
\subsection{Communication Overheads}
Communication overheads is one of the major bottlenecks in federated learning. Existing works try to solve this by either data compression \cite{konevcny2016federated} or by allowing only the relevant outputs by the clients\cite{hsieh2017gaia,luping2019cmfl} to be sent back to the central server.
\subsection{Systems and Data Heterogeneity}
The heterogeneity of the systems in the network as well as the non-identically distributed data from the devices affect the performance of the FL model \cite{mcmahan2017communication,li2020federated}. Although, FedAvg is introduced as a method to tackle the heterogeneity, it is still not robust enough to systems heterogeneity. Works such as \cite{liu2020privacy,bonawitz2019towards} try to address this problem by modifying the model aggregation methods.

\section{Privacy and Security Concerns}
Like any machine learning model, Federated Learning models are also prone to attacks. The attacks can be introduced by a compromised central server or compromised local devices in the learning framework or by any participant in the FL workflow. Attacks in the context of FL will be discussed in this section.
\subsection{Membership Inference Attacks}
Although the raw user data does not leave the local device, there are still many ways to infer the training data used in FL. For instance, in some scenarios, it is possible to infer the information about the training data from the model updates during the learning process. The defense measure looks for mechanism offering a differential privacy guarantee. The most common techniques are  secure computation, differential privacy schemes and running in a trusted execution environment.
\subsubsection{Defense Mechanism 1: Secure Computation}
 Two main techniques come under Secure Computation: Secure Multiparty Computation (SMC) and Homomorphic Encryption. In SMC, two or more parties agree to perform the inputs provided by the participants and reveal the outputs only to a subset of participants. Whereas in homomorphic encryption, computations are performed on  encrypted inputs without decrypting first.
\subsubsection{Defense Mechanism 2: Differential Privacy (DP)}
In differential privacy schemes, the contribution of a user is masked by adding noise to the clipped model parameters before model aggregation \cite{geyer2017differentially}. Some amount of model accuracy will be lost by adding noise to the parameters.
\subsubsection{Defense Mechanism 3:Trusted Execution Environment}
Trusted Execution Environment(TEE) provides a secure platform to run the federated learning process with low computational overhead when compared to secure computation techniques \cite{mo2019efficient}. The current TEE environment is suitable only to CPU devices.
\subsection{Data Poisoning Attacks}
Data Poisoning Attacks are the most common attacks against ML models. In order to launch a data poisoning attack in FL model, adversary poison the training data in a certain number of devices participating in the learning process so that the global model accuracy is compromised. The adversary can poison the data either by directly injecting poisoned data to the targeted device or injecting poisoned data through other devices \cite{sun2020data}. Such attacks can be either targeted or non-targeted. By targeted it means, the adversaries want to influence on the prediction of a subset of classes while deteriorating the global model accuracy\cite{tolpegin2020data}.  
\subsubsection{Defense Mechanisms}
The defense mechanism against such attacks is to identify the malicious participants based on their model updates before model averaging in each round of learning.
\subsection{Model Poisoning Attacks}
Model Poisoning attacks are similar to data poisoning attacks, where the adversary tries to poison the local models instead of the local data. The major motivation behind the model poisoning attack is to introduce errors in the global model. The adversary launches a model poisoning attack by compromising some of the devices and modifying it's local model parameters so that the accuracy of the global model is affected. 
\subsubsection{Defense Mechanisms} The defenses against model poisoning attacks are similar to data poisoning attacks. The most common defense measures are rejections based on Error rate and  Loss function \cite{fang2020local}. In error rate based rejections, the models with significant impact on the error rate on the global model will be rejected. Whereas in loss function based rejections, the models will be rejected based on their impact on loss function of the global model. In some cases, rejections can be made by combining both error based rejections and loss function based rejections.
\subsection{Backdoor Attacks}
Secure averaging in federated learning lets the devices to be anonymous during the model updating process. Using the same functionality, a device or a group of devices can introduce a backdoor functionality in the global model of federated learning \cite{bagdasaryan2020backdoor}. Using a backdoor, an adversary can  mislabel certain tasks without affecting the accuracy of the global model. For instance, an attacker can choose a specific label for a data instance with specific characteristics. Backdoor attacks are also known as targeted attacks. The Intensity of such attacks depends on the proportion of the compromised devices present and model capacity of federated learning \cite{sun2019can} .
\subsubsection{Defense Mechanisms}: The defense against backdoor attacks is either weak differential privacy or norm thresholding of updates. Participant level differential privacy can serve as a form of defense against such attacks but at the cost of performance of the global model \cite{geyer2017differentially}. Whereas norm thresholding can be applied to remove models with boosted model parameters. Even with this defense measures, it is hard to find the malicious participants owing to the secure aggregation techniques and capacity of the deep learning model.
Also FL framework being a distributed system, it might be even harder to manage the randomly misbehaving devices.
\section{Recent Developments in FL}
\subsection{One-shot federated Learning}
In most of the federated learning frameworks, there will be multiple rounds of communication between devices and the central server, which increases the communication overheads. Recently, there is a growing interest in one-shot federated learning which is first introduced by \cite{guha2019one}, where the global model is learned in a single round of communication. In order to overcome communication overheads of sending bulky gradients, \cite{zhou2020distilled} proposes a distilled one-shot federated learning, where each device distills their data and send the fabricated data to the central server. The server then learns the global model by training over the combined data from all the devices.

\subsection{Incentive Mechanisms}
Current FL approaches work under the assumption that devices will cooperate in the learning process whenever required without considering the rewards. Whereas in actual practice, devices or clients must be economically compensated for their participation. To encourage/improve device participation in FL, works such as \cite{kang2019incentive,zhan2020learning,khan2020federated,yu2020sustainable} propose a reputation based incentive mechanism i.e, devices get rewards based on their model accuracy, data reliability and contribution to the global model. However, these works did not talk about how to model convergence and additional communication overheads induced into the framework. 

\subsection{Federated Learning as a Service}
Machine Learning as a Service is getting popular these days and most of them offer only centralized services. In order to offer Federated Learning as a cloud service, it should consider collaboration among third party applications. A recent work \cite{kourtellis2020flaas} tried to develop a FL framework (as a service) which allows applications of third parties to contribute and collaborate on a ML model. The framework is claimed to be suitable for any operation environment as well.

\subsection{Asynchronous Federated Learning}
Most of the current FL aggregation techniques are designed for devices working in a synchronized manner. However, due to systems and data heterogeneity, training and model transfer occur in a asynchronized manner. Therefore it might not be feasible to scale federated optimization in an synchronized manner \cite{xie2019asynchronous}. Works such as \cite{sprague2018asynchronous,van2020asynchronous,xie2019asynchronous} talks about carrying out federated learning in an asynchronous environment. Compared to FedAvg (working in a synchronized manner), asynchronous Federated averaging techniques can handle more devices and allows updates to come at a time.

\subsection{Blockchain in FL}
\begin{figure}
    \centering
    \includegraphics[width=0.42\textwidth]{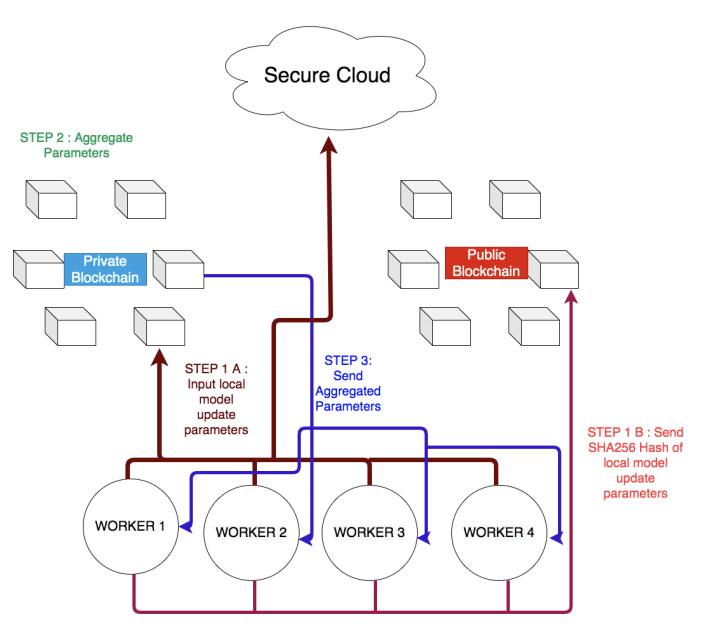}


  \caption{A sample architecture for federated learning over blockchain.}
  \label{block}
\end{figure}
An aggregator is necessary to update the global model managing the asynchronous arrival of parameters from the devices. This can be a constraint for the widespread adoption for the FL models. As blockchain is a decentralized network, devices can collaboratively learn without the central aggregator.  Works such as \cite{ramanan2019baffle,kumar2020blockchain,desai2020blockfla,bao2019flchain} propose Federated Learning in a block-chain framework. A sample architecture \cite{desai2020blockfla} is shown in fig \ref{block}.
\section{Final Remarks }
Federated Learning offers a secure collaborative machine learning framework for different devices without sharing their private data. This attracted a lot of researchers and there is extensive research happening in this domain. Federated Learning has been applied in several domains such as healthcare, transportation etc. Although FL frameworks offer a better privacy guarantee than other ML frameworks, it is still prone to several attacks. The distributed nature of the framework makes it even harder to deploy defense measures. For instance,the gaussian noise added to the local models (for DP) can confuse the aggregation schemes and may result in leaving out the benign participants (while applying model poisoning defense measures). So therefore an interesting research question to pursue will be:
Is it possible to develop a byzantine tolerant FL model while ensuring user privacy using schemes with low computational cost?

\begin{acks}
I would like to thank my course advisor, Dr. Amir Houmansadr for his invaluable guidance and support.
\end{acks}

\bibliographystyle{ACM-Reference-Format}
\bibliography{sample-base}


\end{document}